\pdfoutput=1

\documentclass[11pt]{article}

\usepackage[preprint]{acl}

\usepackage{times}
\usepackage{latexsym}
\usepackage{xcolor}

\usepackage[utf8]{inputenc}

\usepackage{microtype}

\usepackage{inconsolata}

\usepackage[T1]{fontenc}    
\usepackage{url}            
\usepackage{booktabs}       
\usepackage{amsfonts}       
\usepackage{array}
\usepackage{longtable}
\usepackage{wrapfig}
\usepackage{lscape}
\usepackage{adjustbox}
\usepackage{multirow}
\usepackage{tcolorbox}
\usepackage{pgfplots}
\usepackage{pgfplotstable}
\pgfplotsset{compat=1.18}
\usepackage[normalem]{ulem}
\useunder{\uline}{\ul}{}
\usepackage{amsmath}
\usepackage{enumitem}
\usepackage{cleveref}
\usepackage[symbol]{footmisc}
\usepackage{fvextra}
\usepackage{listings}
\usepackage{graphicx}
\usepackage{algorithm}
\usepackage{algpseudocode}
\usepackage{amssymb}

\definecolor{darkgreen}{RGB}{0,100,0}  
\definecolor{darkred}{RGB}{139,0,0}    

\lstdefinestyle{wrappedverbatim}{
  basicstyle=\small\ttfamily,
  breaklines=true,
  breakatwhitespace=false,
  columns=flexible,
  keepspaces=true,
  showstringspaces=false,
  frame=none,
  backgroundcolor=\color{white!95!gray}
}

\title{RAS: Reflection-Augmented Scaling with In-Context Learning for Executable Cypher Query Generation}

\author{
    Minseok Jung$^{1,*}$, Abhas Ricky$^{1}$, Muhammad Rameez Chatni$^{1}$\\[0.5em]
    $^{1}$Cloudera\\
    6220 America Center Dr. \\
    San Jose, CA 95002\\[0.5em]
    \texttt{\{minseok.jung, abhas, mchatni\}@cloudera.com}
}

\begin{document}
\maketitle

\renewcommand{\thefootnote}{}
\renewcommand{\thefootnote}{\arabic{footnote}}

\begin{abstract}
Inference-time scaling can reduce errors in structured query generation, but methods to allocate the compute for query code generation remains underexplored. We study Text2Cypher, where language models generate Cypher queries that execute against property graph databases. Non-executable queries constitute a distinct syntactic failure separate from semantic inaccuracy: a syntax error triggers a system-generated error message from the database. These error messages are typically discarded at inference time rather than leveraged through in-context learning (ICL). We compare two inference methods: Independent Scaling (IS), which performs memoryless resampling, and Reflection-Augmented Scaling (RAS), which conditions each new attempt on prior execution feedback via ICL. Across three Neo4j datasets and five code-specialized language models, RAS reduces the Query Execution Error Rate by 41--50\% at $n{=}5$, outperforming IS at 32--38\%. Execution errors are not merely failures to discard but actionable feedback, and structuring inference-time compute around them is a more efficient path to executability than scaling independent samples.
\end{abstract}

\section{Introduction}

Recent work has shown that allocating additional compute at inference time can substantially improve language model performance on complex reasoning and generation tasks \citealp{snell2024scaling, brown2024monkeys}. Techniques such as scaling self-consistency \citep{wang2023selfconsistency} and repeated sampling \citep{brown2024monkeys} increase the probability of producing better outputs without modifying model parameters \citep{wu2024inference}.

For an efficient scaling, a complementary line of work studies feedback-driven refinement, where models improve outputs using signals from prior attempts. Methods such as Self-Refine \citep{madaan2023selfrefine}, Reflexion \citep{shinn2023reflexion}, and self-debugging\citep{chen2023selfdebug} show that self-critique or execution feedback can guide models toward better generation.

\begin{figure}
    \centering
    \includegraphics[width=\linewidth]{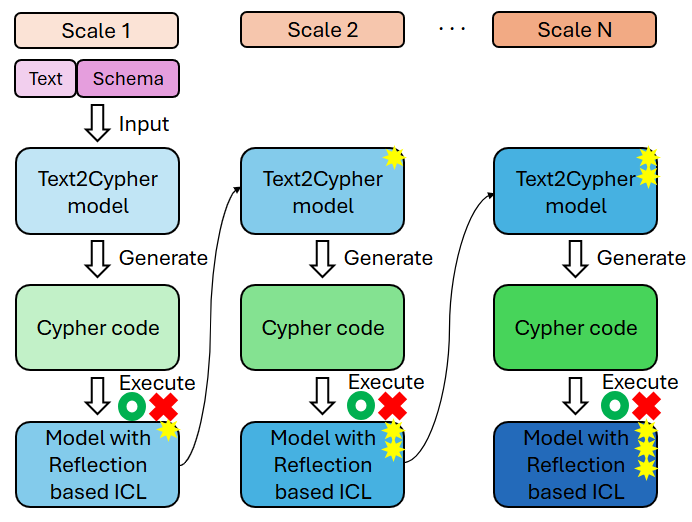}
    \caption{\footnotesize{Our method iteratively generates queries and executes them against the database. When execution fails, the system incorporates execution feedback through reflection-based in-context learning (ICL) to refine subsequent generations. Increasing the inference scale expands the reflection context and improves the probability of producing executable queries}}
    \vspace{-0.9mm}
    \label{fig:enter-label}
\end{figure}

\begin{figure*}[t]
    \centering
    \includegraphics[width=\textwidth]{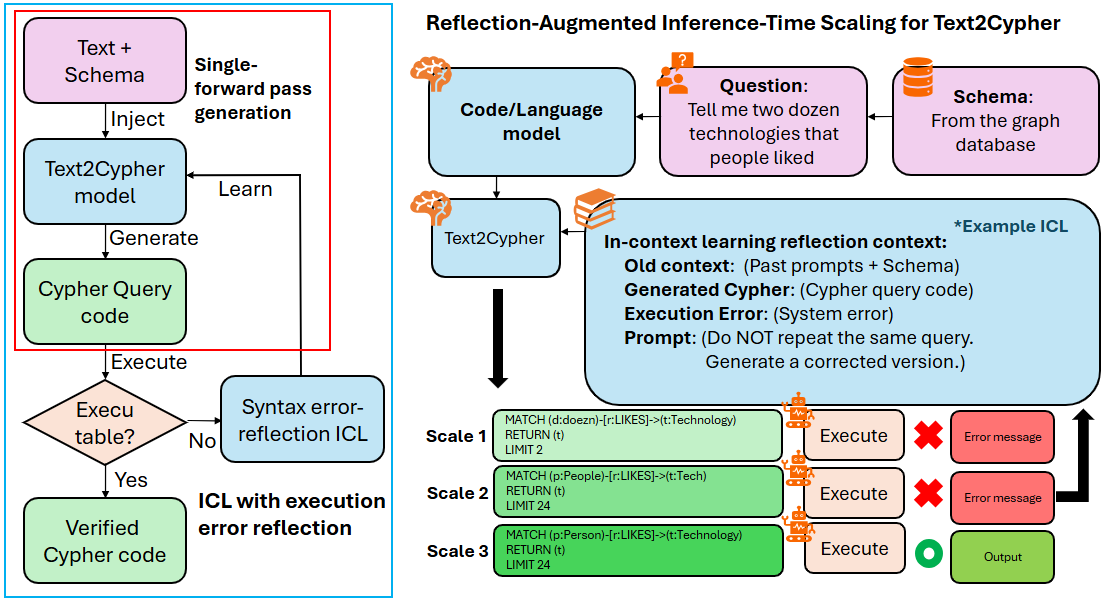}
    \vspace{-2mm}
    \caption{\footnotesize{\textbf{Reflection-augmented inference-time scaling for Text2Cypher.}
    Syntax and schema errors surface to users as execution failures that break interactive query workflows, making them a distinct and user-visible failure mode.
    \textbf{Left:} under single-pass generation, a Cypher query is produced and executed against the graph database; execution failures trigger execution-aware reflection via in-context learning.
    \textbf{Right:} as the inference scale increases, the framework iteratively refines query code by conditioning on feedback from prior failures to reduce syntax errors.}}
    \vspace{-2mm}
    \label{fig:fig2}
\end{figure*}

We study this question in structured generation, where outputs must satisfy strict syntactic and schema constraints. In query code generation, models frequently produce invalid code due to schema mismatches and compositional complexity \citep{gao2023dail, liang2021querying, hains2023natural}. Scaling methods have been invited \citep{shinn2023reflexion, chen2023selfdebug} to overcome this limitation but they did not focus on the executability, which is prerequisite for the accuracy. We introduce \textbf{Reflection-Augmented Scaling (RAS)}, which conditions each new attempt on execution feedback from prior failures, and compare two inference-time strategies for Text2Cypher: \textbf{Independent Scaling (IS)}, which repeatedly samples without memory, and \textbf{RAS}.

Across three Neo4j datasets and five code-specialized language models, RAS consistently outperforms IS under the same compute budget. At $n=5$, RAS achieves an 41--50\% in QER, compared to 32--38\% for IS. Our results show that the effectiveness of inference-time scaling depends not only on how much compute is used, but on how inference strategy is integrated.

\begin{figure*}[t!]
    \centering
    \includegraphics[width=0.99\linewidth]{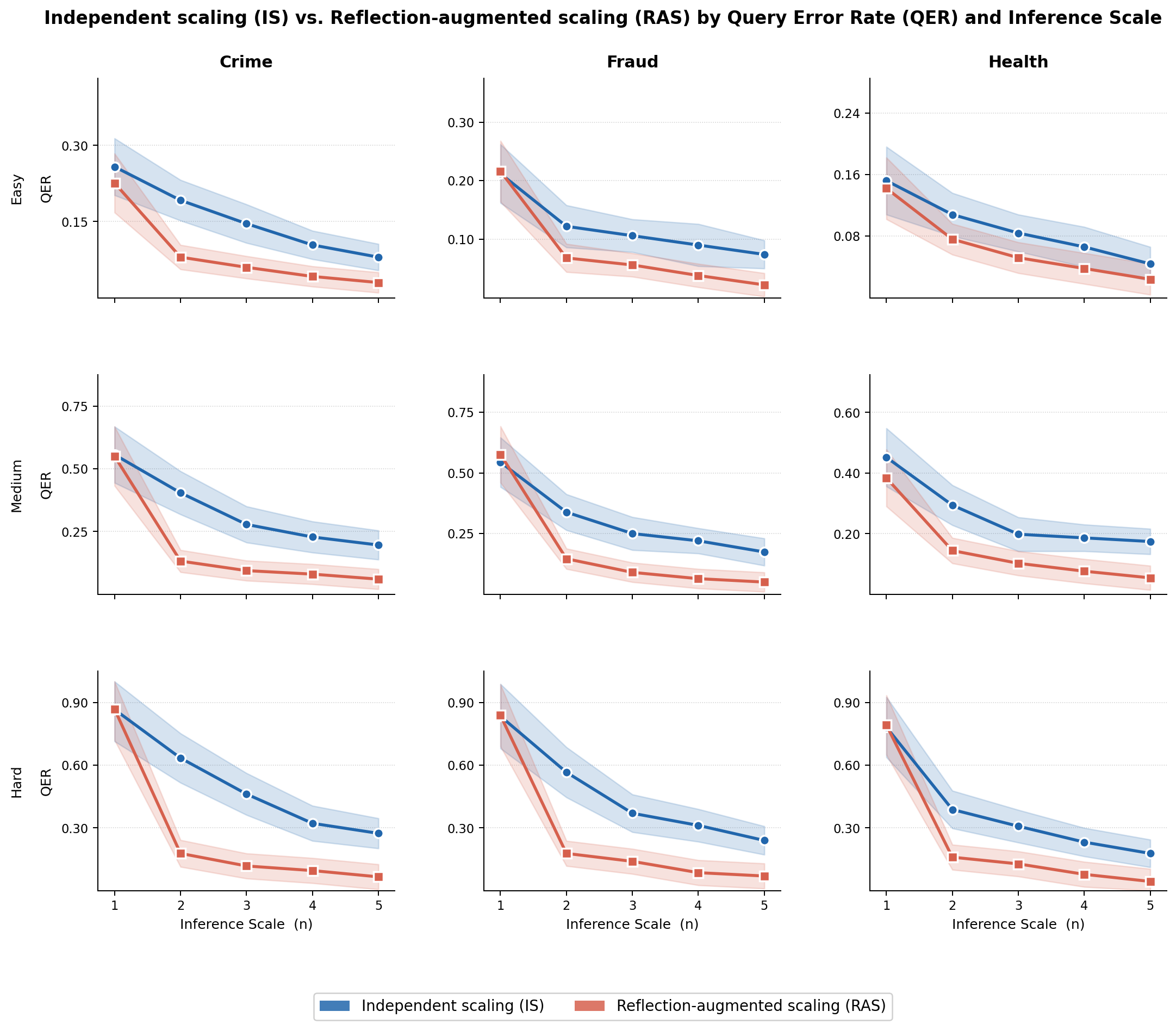}\vspace{-3mm}
    \caption{
    Query Error Rate (QER) across three graph datasets under inference-time scaling. 
    Q@1 denotes baseline single-pass error ($n{=}1$), while IS@5 and RAS@5 report error after scaling to $n{=}5$ using simple re-run and reflection-augmented strategies. Reflection consistently achieves larger absolute reductions in QER. Values are averaged across three query complexities.
    }
    \label{fig:complete_curve}
    \vspace{-3mm}
\end{figure*}

\section{Related Work} 

\paragraph{Semantic parsing for SQL.} Translating natural language into executable database queries has long been studied under the framework of semantic parsing~\citep{berant2013semantic, yih2016value}. In relational settings, the Text2SQL task has benefited from standardized benchmarks such as Spider~\cite{yu2018spider}, enabling systematic evaluation and rapid progress with neural and large language model–based approaches~\cite{wang2020ratsql, scholak2021picard, gao2023dail, pourreza2023din}. Recent work further explores the use of language models as general interfaces to structured databases through in-context learning and agentic workflows~\cite{dong2023c3}. Despite strong performance on relational schemas, these approaches do not transfer directly to graph query languages, which impose fundamentally different structural constraints. 

\paragraph{Semantic parsing for Cypher.} Semantic parsing for graph databases remains comparatively underexplored. Graph query languages such as Cypher require explicit reasoning over heterogeneous node and edge properties, direction-sensitive relationships, and compositional multi-hop traversal patterns. Compared with SQL, graph queries introduce additional structural constraints including path pattern matching and schema sparsity, which increase the likelihood of syntax and execution errors. Prior studies investigate natural-language access to knowledge graphs via SPARQL generation~\cite{lan2022kgqa}, and more recent work explores graph-aware retrieval and fine-tuning approaches for property graph databases~\cite{clemedtson2025graphraft, sivasubramaniam2024sm3}. Nevertheless, the dominant inference paradigm remains single forward-pass generation, without leveraging additional inference-time computation to recover from execution failures. 

\paragraph{Inference-time scaling via independent resampling.} An alternative to improving single-pass generation is to allocate additional inference-time computation through repeated sampling. Test-time compute scaling increases the effective reasoning budget without modifying model parameters, and recent studies show that such strategies can yield substantial gains on complex reasoning tasks~\citep{snell2024scaling, brown2024monkeys, wu2024inference}. Self-consistency~\citep{wang2023selfconsistency} demonstrates that sampling multiple reasoning paths and selecting by majority vote consistently outperforms greedy decoding. In structured query generation, repeated sampling reduces error rates by exploiting stochastic variation across independent outputs. However, independent resampling maintains a fixed conditioning distribution over the input question and database schema, without incorporating evidence from prior execution failures. Consequently, repeated samples tend to reproduce similar structural error modes — such as schema misalignment, incorrect relationship directionality, or invalid aggregation constructs — limiting the gains achievable through scaling alone. 

\paragraph{Execution-grounded reflection and iterative refinement.} A more targeted strategy is to condition subsequent generations on feedback from earlier failures, enabling iterative correction rather than independent exploration. Self-Refine~\cite{madaan2023selfrefine} and Reflexion~\cite{shinn2023reflexion} propose self-refinement loops in which models critique and revise their own outputs, or maintain episodic memory of unsuccessful attempts in agentic reasoning settings. Self-Debugging~\cite{chen2023selfdebug} demonstrates that program execution feedback can guide iterative repair on Text2SQL tasks, improving performance particularly on harder queries. While these methods highlight the value of feedback-driven reasoning, they have not been systematically studied in the context of schema-constrained Cypher generation against live property graph databases. Furthermore, prior work does not directly compare independent resampling and feedback-driven refinement under controlled inference-time compute budgets. Our work addresses these gaps by formalizing execution-aware reflection for Text2Cypher generation and providing a systematic empirical comparison of both strategies across diverse property graph datasets and query complexity levels.

\section{Problem Formulation and Modeling}

\subsection{Text2Cypher as Structured Graph Query Generation}

We formulate Text2Cypher as a structured semantic parsing task that maps a natural 
language question to an executable Cypher query. Let $x \in \mathcal{X}$ denote a 
natural language input and $\mathcal{S}$ denote the database schema. The objective 
is to generate a query $q \in \mathcal{Q}$, where $\mathcal{Q}$ denotes the space 
of syntactically valid Cypher queries over a database instance $G$ consistent 
with $\mathcal{S}$.

We model query generation as conditional structured generation ~\cite{scholak2021picard, wang2020ratsql}:
\begin{equation}
    q \sim p_\theta(q \mid x, \mathcal{S}),
\end{equation}
where $p_\theta$ denotes the conditional distribution induced by an autoregressive 
language model with parameters $\theta$, defining a probability measure over token 
sequences representing candidate Cypher programs. Syntactic validity and 
executability are evaluated only after full decoding. Unlike unconstrained text 
generation, Cypher queries must satisfy strict structural constraints imposed by the 
graph schema; small deviations — such as incorrect relationship directionality or 
references to invalid node properties — result in immediate execution failure.

\subsection{Executability and Accuracy}\label{executability_and_accuracy}

Executability and accuracy measure two conceptually distinct aspects of a generated 
query. \emph{Executability} refers to whether a query can be successfully processed 
by the database engine without raising syntax or schema errors. \emph{Accuracy}, in 
contrast, measures semantic correctness — whether the execution result matches the 
ground-truth answer.

A query may be executable yet semantically incorrect (e.g., due to a missing filter 
or an incorrect aggregation). However, a non-executable query cannot be accurate, 
since it fails before producing any result. This structural asymmetry motivates 
treating execution failure as a distinct and separable source of error.

We formalize this distinction via two binary random variables defined for a generated 
query:
\[
E \in \{0,1\} \quad \text{(syntax)}, \quad A \in \{0,1\} \quad \text{(semantic)}
\]
where $E=1$ denotes a successfully executed query and $E=0$ denotes execution failure due to syntactic or schema violations (e.g., undefined node labels, invalid relationship types, or malformed path syntax), while $A=1$ denotes a result matching the ground-truth answer and $A=0$ denotes a semantic error. Crucially, $E$ captures \emph{structural validity} and $A$ captures \emph{semantic fidelity}: a query may satisfy $E=1$ while failing $A=1$, but $E=0$ guarantees $A=0$ since no result is produced.

We define the joint probabilities as
\[
P_{ij} = \mathbb{P}(E=i,\, A=j),
\]
summarized in Table~\ref{tab:exec_acc_binary}. Since a non-executable query cannot be accurate, $P_{01}$ is undefined. The execution failure probability is therefore $\mathbb{P}(E=0) = P_{00}$, and this is the signal used to trigger the reflection 
loop in Algorithm~\ref{alg:reflection}.  The objective of the RAS algorithm is to decrease $P_{00}$, thereby increasing the total executable mass $P(E=1)=P_{11}+P_{10}$. This paper focuses on reducing execution failure rather than directly optimizing the allocation between semantically correct executions ($P_{11}$) and semantically incorrect but executable queries ($P_{10}$).

\begin{table}[t]
\centering
\small
\begin{tabular}{lcc}
\toprule
 & $A=1$ & $A=0$ \\
\midrule
$E=1$ & $P_{11}$ & $P_{10}$ \\
$E=0$ & \text{---} & $P_{00}$ \\
\bottomrule
\end{tabular}
\vspace{-2mm}
\caption{\footnotesize Joint distribution of executability ($E$) and semantic 
accuracy ($A$). The cell $P_{01}$ is undefined since a non-executable query 
cannot produce a correct result.}
\vspace{-4mm}
\label{tab:exec_acc_binary}
\end{table}

\subsection{Query Execution Error}\label{modeling:qer}

Given a generated query $q$ and a fixed database instance $G$, execution is 
deterministic. We write
\begin{equation}
    (r,\, m) = \mathcal{E}(q, G),
\end{equation}
where $r$ is the query result returned by the database engine and $m$ is the 
accompanying execution status message (e.g., a success confirmation or a structured 
error diagnostic). Accuracy $A$ is evaluated against $r$, while executability $E$ 
is determined by $m$: a query fails ($E=0$) if and only if $m$ indicates an error, 
in which case $r$ is undefined.

We define the \emph{Query Execution Error} (QEE) indicator as
\begin{equation}
    \mathrm{QEE}(q, G)
    \;:=\;
    \mathbf{1}\!\left[m \in \mathcal{M}_{\mathrm{error}}\right],
\end{equation}
where $\mathcal{M}_{\mathrm{error}}$ denotes the set of error-class status messages. 
Given an input distribution $\mathcal{D}$ over $x$, the \emph{Query Execution Error 
Rate} (QER) is
\begin{equation}
    \mathrm{QER}
    \;:=\;
    \mathbb{E}_{x \sim \mathcal{D}}
    \;
    \mathbb{E}_{q \sim p_\theta(\cdot \mid x,\mathcal{S})}
    \!\left[
        \mathrm{QEE}(q, G)
    \right].
\end{equation}
Thus $\mathrm{QER}$ equals $\mathbb{P}(E=0)$ and is estimated empirically as the 
fraction of queries that fail execution across all inputs.

From a system perspective, executability acts as a hard constraint on downstream correctness: when E = 0, no result is produced, and the retrieval stage in RAG pipelines fails entirely. Consequently, reducing P(E = 0) is not only a syntactic objective but a prerequisite for enabling semantically grounded responses.

\subsection{Inference-Time Scaling Strategies}

We study two inference-time strategies under a fixed compute budget $T$: 
(i) \emph{Independent Scaling} (IS) and 
(ii) \emph{Reflection-Augmented Scaling} (RAS).
Both allocate additional test-time computation without updating model parameters.

\paragraph{Independent Scaling (IS).}
IS repeatedly samples from the same conditional distribution 
$p_\theta(\cdot \mid x,\mathcal{S})$ until an executable query is obtained or the 
budget $T$ is exhausted, as formalized in Algorithm~\ref{alg:rerun}. Each sample is 
generated independently, without incorporating information from prior execution 
failures\cite{wang2023selfconsistency, brown2024monkeys}. If no executable query is found within $T$ attempts, the final sampled 
query is returned as a fallback.

\begin{algorithm}[t]
\small
\caption{Independent Scaling (IS)}
\label{alg:rerun}
\begin{algorithmic}[1]
\Require Input $x$, schema $\mathcal{S}$, graph $G$, budget $T$
\State $q_{\text{last}} \gets \varnothing$
\For{$t = 1$ \textbf{to} $T$}
    \State $q_t \sim p_\theta(\cdot \mid x, \mathcal{S})$
    \State $q_{\text{last}} \gets q_t$
    \State $m_t \gets \mathcal{E}(q_t, G)$
    \If{$m_t \notin \mathcal{M}_{\mathrm{error}}$}
        \State \Return $q_t$
    \EndIf
\EndFor
\State \Return $q_{\text{last}}$
\end{algorithmic}
\end{algorithm}

\paragraph{Reflection-Augmented Scaling (RAS).}
RAS augments the generation context with execution feedback from prior failed 
attempts, enabling iterative targeted correction ~\cite{madaan2023selfrefine, shinn2023reflexion}. After each execution failure, the 
failed query and its error message are appended to the conditioning context 
$\mathcal{C}$. Subsequent generations condition on this expanded context, allowing 
the model to identify and correct specific structural errors. Algorithm~\ref{alg:reflection} 
formalizes this procedure. As with IS, if no executable query is produced within 
$T$ attempts, the final generated query is returned as a fallback.

\begin{algorithm}[t]
\small
\caption{Reflection-Augmented Scaling (RAS)}
\label{alg:reflection}
\begin{algorithmic}[1]
\Require Input $x$, schema $\mathcal{S}$, graph $G$, budget $T$
\State $\mathcal{C} \gets (x, \mathcal{S})$
\State $q_{\text{last}} \gets \varnothing$
\For{$t = 1$ \textbf{to} $T$}
    \State $q_t \sim p_\theta(\cdot \mid \mathcal{C})$
    \State $q_{\text{last}} \gets q_t$
    \State $m_t \gets \mathcal{E}(q_t, G)$
    \If{$m_t \notin \mathcal{M}_{\mathrm{error}}$}
        \State \Return $q_t$
    \EndIf
    \State $\mathcal{C} \gets \mathcal{C} \cdot (q_t, m_t)$  \Comment{Append failure to context}
\EndFor
\State \Return $q_{\text{last}}$
\end{algorithmic}
\end{algorithm}

\section{Experimental Setup}
\label{sec:experimental_setup}

\subsection{Datasets}
\begin{table}[t]
\small
\centering
\begin{tabular}{lccc}
\toprule
\textbf{Dataset} & \textbf{Nodes} & \textbf{Edges} & \textbf{Domain} \\
\midrule
Healthcare  & 11,381     & 61,453     & Clinical Records \\
Fraud       & 332,973    & 980,098    & Financial Transactions \\
Crime       & 61,521     & 105,840    & Public Safety \\
\bottomrule
\end{tabular}
\vspace{-2mm}
\caption{\footnotesize Statistics of the Neo4j property-graph datasets used in our Text2Cypher evaluation. The datasets vary substantially in scale and structural density, enabling experiments across heterogeneous schema structures.}
\vspace{-6mm}
\label{tab:neo4j_datasets}
\end{table}

We evaluate our approach on three publicly available Neo4j property graph databases spanning diverse scales and structural characteristics: \textit{Healthcare}, \textit{Fraud}, and \textit{Crime} (Table~\ref{tab:neo4j_datasets}). These datasets vary substantially in graph size, schema complexity, and query topology, enabling robustness assessment across heterogeneous graph structures.

\textbf{Healthcare} is derived from FDA adverse event reporting data and contains heterogeneous biomedical entities including drugs, patients, and reported reactions. Queries over this graph frequently require multi-entity joins and attribute-based filtering, stressing schema alignment and type consistency during generation. \textbf{Fraud} represents a large-scale synthetic financial transaction network characterized by high connectivity and long relational paths. The dense topology increases the likelihood of structural errors in generated Cypher queries, particularly in aggregation and multi-hop traversal patterns. \textbf{Crime} follows the investigative POLE (Person--Object--Location--Event) data model used in law-enforcement analytics, exhibiting moderate scale but complex relationship directionality and path constraints.

\subsection{Scope of Evaluation}
This experiment evaluates execution reliability rather than full semantic correctness. Our metric, QER, measures whether a generated Cypher query can be executed successfully against the target GDB. This choice is deliberate: in graph-based RAG and agentic database systems, non-executable queries represent a hard failure mode because no evidence can be retrieved for downstream reasoning before measuring accuracy. To put it another way, executability is a necessary condition for accuracy despite it is not a sufficient condition. Although query code executability is not a substitute for semantic accuracy, it is a required precondition.

\subsection{Question Generation}

For each dataset, natural language questions are generated at three complexity levels (\textit{Easy}, \textit{Medium}, and \textit{Hard}), conditioned on the graph schema and representative subgraph samples. Full prompt templates and representative examples are provided in the Appendix.

\subsection{Text2Cypher}
We implement Text2Cypher using LangChain's \texttt{GraphCypherQAChain}, which constructs a structured generation prompt by injecting the database schema $\mathcal{S}$ retrieved directly from the Neo4j instance. To isolate the query generation stage and focus directly on QEE, we do not evaluate downstream answer for accuracy. Each generated query is executed against the live database instance to determine executability as defined in §\ref{executability_and_accuracy}.

\subsection{Models}
We evaluate five open-weight, code-specialized language models: \texttt{CodeLlama-7B} and \texttt{CodeLlama-13B}~\cite{roziere2023codellama}, \texttt{DeepSeek-Coder-6.7B}~\cite{guo2024deepseek}, \texttt{Qwen2.5-Coder-7B}~\cite{hui2024qwen25coder}, and \texttt{StarCoder2-7B}~\cite{lozhkov2024starcoder2}. These models are pretrained on large-scale source code corpora, inducing a structural prior toward syntactically well-formed token sequences. We deliberately focus on small language models (SLMs) to reduce the latency of the inference stage. All models are evaluated under identical schema injection, decoding configuration, and execution-validation settings within the same \texttt{GraphCypherQAChain} pipeline.

\subsection{Scaled Context Construction via In-Context Learning}

We instantiate the two inference methods (Algorithms~\ref{alg:rerun} and~\ref{alg:reflection}) at scale. For our method, Alg. \ref{alg:reflection}, the initial context $\mathcal{C}_0 = (x, \mathcal{S})$ is constructed by injecting the schema $\mathcal{S}$ directly from the Neo4j instance into a structured Cypher generation prompt. For the RAS, each failed attempt appends the generated query $q_t$ and its execution error $m_t$ to the prompt, along with a corrective instruction, forming $\mathcal{C}_t = \mathcal{C}_{t-1} \cup \{q_t, m_t\}$. This accumulated context is embedded into the prompt, preserving compatibility with the system architecture. Alg. \ref{alg:rerun} keeps generating answers until it succeeds. At $T{=}1$, both strategies are based on the identical single-pass baseline, but the scaling with RAS keeps adding query code with system errors. 

\subsection{Hyperparameters}

Each experimental configuration — defined by a dataset, model, complexity level, and inference strategy — is evaluated over 128 independent runs. We report the mean QER along with its standard deviation $\sigma_{\mathrm{QER}}$ across runs. All decoding is performed at temperature $\tau = 0.9$ to introduce sufficient stochastic diversity across trials while maintaining generation quality~\cite{wang2023selfconsistency}. The inference scale is fixed at $T = 5$, selected via multi-objective optimization (MOO) on the trade-off between error reduction, latency and compute cost with the optimality identified at the knee point of Pareto frontiers \citep{jung2025moo}.

Evaluation is based on the query execution that has been introduced in Table \ref{executability_and_accuracy}. Additional information about the prompts by complexity and results by complexity are provided in Appendix \ref{reseults_details} and Appendix \ref{input_complexities}.

\begin{table*}[t]
\centering
\scriptsize
\setlength{\tabcolsep}{4.2pt}
\renewcommand{\arraystretch}{1.02}
\begin{tabular}{lccccc|ccccc|ccccc}
\toprule
& \multicolumn{5}{c}{\textbf{Crime}}
& \multicolumn{5}{c}{\textbf{Fraud}}
& \multicolumn{5}{c}{\textbf{Healthcare}} \\
\cmidrule(lr){2-6} \cmidrule(lr){7-11} \cmidrule(lr){12-16}
\textbf{Model}
& Q@1 & IS@5 & RAS@5 & $\Delta$IS & $\Delta$RAS
& Q@1 & IS@5 & RAS@5 & $\Delta$IS & $\Delta$RAS
& Q@1 & IS@5 & RAS@5 & $\Delta$IS & $\Delta$RAS \\
\midrule
CodeLlama-13B
& 0.40 & 0.14 & 0.05 & 0.26 & \textbf{0.35}
& 0.38 & 0.12 & 0.03 & 0.26 & \textbf{0.35}
& 0.32 & 0.08 & 0.03 & 0.24 & \textbf{0.29} \\

DeepSeek-6.7B
& 0.46 & 0.18 & 0.04 & 0.28 & \textbf{0.42}
& 0.44 & 0.13 & 0.05 & 0.31 & \textbf{0.39}
& 0.37 & 0.13 & 0.04 & 0.24 & \textbf{0.33} \\

CodeLlama-7B
& 0.58 & 0.19 & 0.05 & 0.39 & \textbf{0.53}
& 0.56 & 0.17 & 0.06 & 0.39 & \textbf{0.50}
& 0.43 & 0.13 & 0.05 & 0.30 & \textbf{0.38} \\

Qwen2.5-7B
& 0.65 & 0.21 & 0.06 & 0.44 & \textbf{0.59}
& 0.63 & 0.19 & 0.04 & 0.44 & \textbf{0.59}
& 0.51 & 0.14 & 0.04 & 0.37 & \textbf{0.47} \\

StarCoder2-7B
& 0.68 & 0.20 & 0.06 & 0.48 & \textbf{0.62}
& 0.69 & 0.20 & 0.05 & 0.49 & \textbf{0.64}
& 0.63 & 0.17 & 0.05 & 0.46 & \textbf{0.58} \\
\bottomrule
\end{tabular}
\caption{
Query Execution Error Rate (QER) across three graph datasets and five coder models. Q@1 denotes baseline error at a single generation ($n{=}1$), computed as the average of IS@1 and RAS@1 across Easy, Medium, and Hard complexity levels. IS@5 and RAS@5 denote error after test-time scaling to $n{=}5$ using IS (Simple Retrials) and RAS (Reflection-Augmented Scaling). $\Delta$IS and $\Delta$RAS represent reductions in QER relative to Q@1. The average values are rounded up into two decimal points. Detailed experimenta result is in Appendix \ref{tab:overall_results}}.
\label{tab:main_results_full}
\end{table*}

\section{Results}

\subsection{Inference Scaling Consistently Reduces QER}
\label{subsec:results_scaling}

Table~\ref{tab:main_results_full} reports QER under single-shot generation ($n{=}1$) and scaled inference ($n{=}5$) for both IS and RAS. Across all datasets and models, increasing the inference scale reduces QER regardless of the inference methods. The Q@1 is the average of the two methods, and both showed a similar error rate at the initial stage because the ICL of RAS is not applicable at RAS@1.

\subsection{RAS Outperforms Independent Scaling}
\label{subsec:results_reflection}

At scale $n{=}1$, both strategies shows similar QER, as no execution feedback has yet been incorporated. However, at $n{=}5$, RAS achieves QER of 3--6\% across most configurations, representing a reduction of more than 10\% over IS. This gap demonstrates that execution feedback constitutes a qualitatively stronger inference-time signal than independent resampling. Rather than simply retrying query generations, RAS actively incorporates information from prior execution failures to steer subsequent generation toward executable queries. The advantage of RAS over IS is consistent across all datasets, model architectures, and input complexities.

\subsection{QER Scaling Trajectories Differ Between Strategies}
\label{subsec:results_trajectory}

Beyond endpoint comparisons, Figure~\ref{fig:complete_curve} reveals a difference in how the two strategies reduce QER across the inference budget. IS exhibits an approximately linear decrease in QER as the scale grows from 1 to 5, suggesting that each additional independent trial contributes a roughly constant marginal gain in executability. RAS, by contrast, follows a concave trajectory: the sharpest drop occurs at $n{=}2$, the first point at which execution feedback from a failed attempt becomes available, after which further improvements taper gradually. This suggests that the primary benefit of execution-grounded reflection is concentrated in the earliest feedback step, while subsequent iterations yield diminishing returns.

\subsection{RAS is Robust Across Datasets, Models, and Query Complexities}
\label{subsec:results_robustness}

A critical requirement for any inference-time strategy is that its gains generalize beyond a narrow set of conditions. Table~\ref{tab:main_results_full} demonstrates that RAS show a greater QER decrease consistently across datasets and models. Across all three datasets — Healthcare, Fraud, and Crime — RAS reduces QER to 0.03--0.06 at $n{=}5$, despite differences are observed by graph topology, schema complexity, and domain vocabulary. Robustness also holds across model architectures. RAS achieves consistent gains over all five code-specialized models spanning. This uniformity suggests that the feedback mechanism is not dependent on any particular model's prior capacity for schema-aligned generation. 
Further, the standard deviation of QER ($\sigma_{\mathrm{QER}}$) and consistency across input complexities reported in Table~\ref{tab:overall_results} provides further evidence of the robustness. Crucially, these values are consistent across model architectures and datasets with no outlying configurations, indicating that RAS produces reliably low error rates across independent runs rather than occasional successes masking high variance. The concrete failure signal incorporated at each reflection step guides generation toward a stable, executable region of the query space, constraining the stochastic variation inherent in autoregressive decoding.

\section{Conclusion}

This paper studies inference-time scaling for Text2Cypher generation by formalizing and comparing two strategies: Independent Scaling (IS) and Reflection-Augmented Scaling (RAS). We evaluate both approaches under a fixed compute budget across three Neo4j property graph datasets, five code-specialized language models, and multiple query complexity levels.

Our results reveal that IS and RAS exhibit fundamentally different scaling behaviors. IS reduces the Query Execution Error Rate (QER) in an approximately linear manner, where each additional sample provides a consistent marginal improvement through stochastic diversity. In contrast, RAS follows a concave trajectory, achieving its largest reduction at $n{=}2$, when execution feedback first becomes available, followed by diminishing returns. This two-phase behavior highlights a key asymmetry: the primary benefit of reflection is concentrated in the earliest feedback-driven correction step rather than in prolonged iterative refinement.

Empirically, RAS consistently outperforms IS under identical compute budgets, achieving substantially lower QER across all datasets and models. Notably, no IS configuration matches the performance of RAS at the same inference scale. These findings demonstrate that the structure of inference-time computation—specifically, whether it incorporates feedback—can be more critical than increasing model size or the number of independent samples.

Overall, our results suggest that effective inference-time scaling for structured generation requires moving beyond independent resampling toward feedback-driven strategies. Execution feedback provides a structured and informative signal that transforms the correction process from stochastic exploration into targeted refinement. Understanding how to optimally allocate compute across such feedback mechanisms has direct implications for deploying reliable and cost-efficient language model systems in graph query generation.

Although our experiments focus on Text2Cypher, the proposed framework is not limited to graph query languages. The same execution-grounded reflection mechanism can be extended to other structured semantic parsing tasks, such as Text2SQL, where outputs must satisfy strict syntactic and schema constraints. More broadly, our findings point toward a general paradigm in which verifiable feedback signals can improve inference-time scaling in program synthesis and agentic systems.

\section{Limitations}
This work has several limitations. First, our evaluation is restricted to Text2Cypher on Neo4j property graphs; inference scaling dynamics may differ for other semantic parsing tasks (e.g., Text2SQL) or database engines with distinct execution semantics.

Second, we cap the inference budget at $n{=}5$, reflecting a practical performance–latency trade-off; larger budgets may yield diminishing returns, higher latency, or over-correction from compounding feedback.

Third, we focus on open-weight, code-specialized models for reproducibility, leaving the scaling behavior of larger proprietary models as an open question.

Fourth, we evaluate performance primarily through query executability. Although executability does not guarantee semantic correctness, we treat it as a prerequisite reliability condition: in graph-based RAG and agentic systems, a non-executable query halts evidence retrieval entirely, causing the system to fail before any answer can be grounded in data. Reducing execution errors is therefore a necessary first step toward reliable downstream reasoning, and future work should extend this framework to jointly optimize executability and semantic fidelity.

Finally, our reliance on in-context learning to incorporate execution feedback introduces token and latency overhead as the reflection context grows. Execution traces are also used transiently rather than persisted; aggregating these logs could enable system-level learning across queries beyond single-session inference.

\section*{Acknowledgment}
Thanks to Nashua Springberry, Michael Schuler, Sreenath Somarajapuram, and Neelabh Pant for constructive comments.

\bibliography{main_latex}

\appendix

\section{Appendix}

\subsection{Experimental Results}\label{reseults_details}
Table~\ref{tab:overall_results} provides a granular breakdown of the Query Execution Error Rate (QER) across all evaluated datasets, models, and query complexities. As expected, single-pass generation (QER@1) degrades significantly as query difficulty increases, with execution failures climbing as high as 0.97 on Hard queries. While scaling to $n{=}5$ via Independent Scaling (IS@5) yields moderate improvements, Reflection-Augmented Scaling (RAS@5) consistently suppresses the error rate to below 0.10, even on the most topologically complex queries. Furthermore, the reported standard deviations ($\sigma_{\text{IS}}$ and $\sigma_{\text{RAS}}$) reveal that RAS not only reduces the mean error but also tightens the variance across independent runs—particularly at the Hard complexity level. This indicates that execution-grounded reflection effectively constrains the generation space, yielding vastly more stable and predictable behavior than simple independent resampling.

\begin{table*}[t]
\centering
\small
\setlength{\tabcolsep}{5pt}
\begin{tabular}{llccccc}
\toprule
\textbf{Model} & \textbf{Cmplx.}
  & \textbf{QER@1} & \textbf{IS@5} & \textbf{RAS@5}
  & $\sigma_{\text{IS}}$ & $\sigma_{\text{RAS}}$ \\
\midrule
\multicolumn{7}{c}{\textsc{Crime}} \\
\midrule
\multirow{3}{*}{CodeLlama-13B}
  & Easy   & 0.17 & 0.04 & 0.02 & 0.02 & 0.02 \\
  & Medium & 0.34 & 0.17 & 0.05 & 0.04 & 0.04 \\
  & Hard   & 0.70 & 0.22 & 0.07 & 0.06 & 0.06 \\
\multirow{3}{*}{DeepSeek-Coder-6.7B}
  & Easy   & 0.21 & 0.06 & 0.02 & 0.03 & 0.02 \\
  & Medium & 0.35 & 0.24 & 0.05 & 0.06 & 0.04 \\
  & Hard   & 0.83 & 0.23 & 0.06 & 0.07 & 0.05 \\
\multirow{3}{*}{CodeLlama-7B}
  & Easy   & 0.21 & 0.08 & 0.04 & 0.03 & 0.02 \\
  & Medium & 0.63 & 0.16 & 0.06 & 0.06 & 0.04 \\
  & Hard   & 0.91 & 0.32 & 0.06 & 0.08 & 0.04 \\
\multirow{3}{*}{Qwen2.5-Coder-7B}
  & Easy   & 0.30 & 0.09 & 0.03 & 0.02 & 0.02 \\
  & Medium & 0.70 & 0.19 & 0.07 & 0.06 & 0.04 \\
  & Hard   & 0.96 & 0.35 & 0.07 & 0.07 & 0.06 \\
\multirow{3}{*}{StarCoder2-7B}
  & Easy   & 0.34 & 0.13 & 0.04 & 0.03 & 0.02 \\
  & Medium & 0.76 & 0.22 & 0.07 & 0.07 & 0.04 \\
  & Hard   & 0.95 & 0.25 & 0.07 & 0.08 & 0.07 \\
\midrule
\multicolumn{7}{c}{\textsc{Fraud}} \\
\midrule
\multirow{3}{*}{CodeLlama-13B}
  & Easy   & 0.12 & 0.05 & 0.01 & 0.02 & 0.02 \\
  & Medium & 0.33 & 0.14 & 0.03 & 0.04 & 0.04 \\
  & Hard   & 0.69 & 0.18 & 0.04 & 0.06 & 0.07 \\
\multirow{3}{*}{DeepSeek-Coder-6.7B}
  & Easy   & 0.14 & 0.05 & 0.02 & 0.03 & 0.02 \\
  & Medium & 0.44 & 0.15 & 0.06 & 0.05 & 0.04 \\
  & Hard   & 0.75 & 0.18 & 0.08 & 0.06 & 0.06 \\
\multirow{3}{*}{CodeLlama-7B}
  & Easy   & 0.18 & 0.09 & 0.03 & 0.02 & 0.02 \\
  & Medium & 0.63 & 0.19 & 0.06 & 0.06 & 0.04 \\
  & Hard   & 0.87 & 0.24 & 0.09 & 0.07 & 0.06 \\
\multirow{3}{*}{Qwen2.5-Coder-7B}
  & Easy   & 0.28 & 0.07 & 0.03 & 0.02 & 0.02 \\
  & Medium & 0.68 & 0.18 & 0.04 & 0.07 & 0.04 \\
  & Hard   & 0.92 & 0.32 & 0.06 & 0.06 & 0.06 \\
\multirow{3}{*}{StarCoder2-7B}
  & Easy   & 0.36 & 0.11 & 0.02 & 0.03 & 0.02 \\
  & Medium & 0.73 & 0.21 & 0.06 & 0.06 & 0.04 \\
  & Hard   & 0.97 & 0.28 & 0.08 & 0.09 & 0.08 \\
\midrule
\multicolumn{7}{c}{\textsc{Health}} \\
\midrule
\multirow{3}{*}{CodeLlama-13B}
  & Easy   & 0.08 & 0.01 & 0.01 & 0.02 & 0.02 \\
  & Medium & 0.24 & 0.13 & 0.04 & 0.04 & 0.04 \\
  & Hard   & 0.63 & 0.11 & 0.05 & 0.06 & 0.05 \\
\multirow{3}{*}{DeepSeek-Coder-6.7B}
  & Easy   & 0.08 & 0.03 & 0.02 & 0.02 & 0.02 \\
  & Medium & 0.32 & 0.15 & 0.06 & 0.04 & 0.03 \\
  & Hard   & 0.71 & 0.22 & 0.03 & 0.07 & 0.06 \\
\multirow{3}{*}{CodeLlama-7B}
  & Easy   & 0.13 & 0.03 & 0.02 & 0.02 & 0.02 \\
  & Medium & 0.38 & 0.17 & 0.06 & 0.05 & 0.04 \\
  & Hard   & 0.79 & 0.19 & 0.06 & 0.07 & 0.04 \\
\multirow{3}{*}{Qwen2.5-Coder-7B}
  & Easy   & 0.20 & 0.07 & 0.03 & 0.03 & 0.02 \\
  & Medium & 0.47 & 0.19 & 0.05 & 0.05 & 0.04 \\
  & Hard   & 0.86 & 0.16 & 0.03 & 0.06 & 0.06 \\
\multirow{3}{*}{StarCoder2-7B}
  & Easy   & 0.25 & 0.08 & 0.04 & 0.03 & 0.02 \\
  & Medium & 0.69 & 0.23 & 0.06 & 0.04 & 0.04 \\
  & Hard   & 0.95 & 0.21 & 0.05 & 0.07 & 0.06 \\
\bottomrule
\end{tabular}
\caption{QER per dataset at $n{=}1$ (QER@1, averaged over IS and RAS baselines) and $n{=}5$ (IS@5, RAS@5).
$\sigma_{\text{IS}}$ and $\sigma_{\text{RAS}}$ denote the standard deviation of QER at $n{=}5$ under IS and RAS, respectively.}
\label{tab:overall_results}
\end{table*}

\subsection{Input text complexities}\label{input_complexities}
Table~\ref{tab:detailed_queries} provides a representative sample of the natural language queries used in our evaluation, categorized by dataset and complexity tier. To provide a concrete understanding of the structural challenges introduced at each level, we detail the specific input prompts alongside their corresponding evaluation identifiers. As illustrated in the table, \textit{Easy} queries (e.g., \texttt{crime\_easy\_1}) focus on straightforward entity retrieval and direct attribute filtering. \textit{Medium} queries (e.g., \texttt{health\_medium\_1}) introduce relational constraints, requiring the model to successfully join multiple entity types. Finally, \textit{Hard} queries (e.g., \texttt{fraud\_hard\_1}) demand advanced structural reasoning, requiring the generated Cypher code to handle cyclic multi-hop paths, mutual relationships, and conditional aggregations simultaneously. These examples highlight the escalating topological complexity that the inference strategies must navigate.

\begin{table*}[ht!]
\centering
\small
\renewcommand{\arraystretch}{1.3}
\begin{tabular}{@{}llp{10.5cm}@{}}
\toprule
\textbf{Query ID} & \textbf{Complexity} & \textbf{Natural Language Question} \\
\midrule
\multicolumn{3}{@{}l}{\textbf{Crime Dataset}} \\
\midrule
\texttt{crime\_easy} & Easy & Find all crime incidents and return their incident IDs and crime types. \\
\texttt{crime\_medium} & Medium & Find people involved in crime incidents at a specific location and return the person name and incident ID. \\
\texttt{crime\_hard} & Hard & Identify repeat offenders who are connected to more than one crime incident and return the person name and the number of incidents. \\
\midrule
\multicolumn{3}{@{}l}{\textbf{Fraud Dataset}} \\
\midrule
\texttt{fraud\_easy} & Easy & List all transactions that are marked as fraudulent and return their transaction IDs. \\
\texttt{fraud\_medium} & Medium & Find customers who have more than two fraudulent transactions and return the customer name and the number of fraudulent transactions. \\
\texttt{fraud\_hard} & Hard & Find pairs of accounts that have transferred money between each other and are both involved in at least one fraudulent transaction, and return the account IDs and the number of shared fraudulent transactions. \\
\midrule
\multicolumn{3}{@{}l}{\textbf{Healthcare Dataset}} \\
\midrule
\texttt{health\_easy} & Easy & Find all patients in the database and return their names. \\
\texttt{health\_medium} & Medium & List all medical records for a specific patient and the physicians associated with those records. \\
\texttt{health\_hard} & Hard & Find patients who have been treated by more than one physician for the same medical condition and return the condition and physician count. \\
\bottomrule
\end{tabular}
\vspace{2mm}
\caption{Detailed natural language questions across the Crime, Fraud, and Healthcare datasets, including their assigned evaluation identifiers and complexity tiers.}
\label{tab:detailed_queries}
\end{table*}

\section{Broader Applications: Reflection as a Generalized Reasoning Engine}
\label{app:reasoning_engine}

While this study focuses on Text2Cypher as a rigorous testbed for execution-grounded scaling, the underlying reflection-augmented in-context learning (ICL) mechanism is fundamentally domain-agnostic. Beyond structured query generation, this iterative feedback loop can serve as the core component of a generalized advanced reasoning engine.

In standard language model deployments, complex reasoning is often bottlenecked by a single-forward pass. By treating environmental feedback—such as database execution errors, compiler outputs, or external API responses—as structured diagnostic signals, the Reflection-Augmented Scaling (RAS) methodology effectively transforms a static generation task into an agentic, self-correcting workflow. 

As a generalized reasoning engine, this architecture can be extended to multi-turn analytical tasks where a model must autonomously navigate knowledge graphs, verify its own intermediate logical steps, and recover from hallucinations before returning a final answer to the user. Consequently, understanding the scaling trajectory of reflection (e.g., the sharp performance gains at $n{=}2$) informs the compute-optimal design of these broader agentic systems, ensuring that inference-time compute is allocated efficiently during complex, multi-step reasoning tasks.

The full production implementation is not publicly released because it is part of the corporation's internal reasoning-engine. To support reproducibility, we describe the inference procedure, prompt construction, execution-validation protocol, datasets, model choices, decoding settings, and evaluation metrics in detail. We also provide pseudocode for both Independent Scaling and Reflection-Augmented Scaling, enabling independent reimplementation of the experimental logic without exposing proprietary system components.
\end{document}